\documentclass{llncs}

\usepackage{makeidx}  
\usepackage{amsmath}
\usepackage{amstext}
\usepackage{amssymb}
\usepackage{graphicx}
\usepackage{wrapfig}

\begin{document}

\frontmatter          

\pagestyle{headings}  
\addtocmark{Patch-Based Image Segmentation} 

\mainmatter              

\title{A Latent Source Model for \\ Patch-Based Image Segmentation}

\titlerunning{Patch-Based Image Segmentation}

\author{George H.~Chen \and Devavrat Shah \and Polina Golland}

\authorrunning{G.~H.~Chen et al.} 

\tocauthor{George H.~Chen, Devavrat Shah, and Polina Golland}

\institute{Massachusetts Institute of Technology, Cambridge MA 02139, USA}

\maketitle

\begin{abstract}
Despite the popularity and empirical success of patch-based nearest-neighbor
and weighted majority voting approaches to medical image segmentation, there
has been no theoretical development on when, why, and how well these
nonparametric methods work. We bridge this gap by providing a theoretical
performance guarantee for nearest-neighbor and weighted majority voting
segmentation under a new probabilistic model for patch-based image
segmentation. Our analysis relies on a new local property for how similar
nearby patches are, and fuses existing lines of work on modeling natural
imagery patches and theory for nonparametric classification. We use the model
to derive a new patch-based segmentation algorithm that iterates between
inferring local label patches and merging these local segmentations to produce
a globally consistent image segmentation. Many existing patch-based algorithms
arise as special cases of the new algorithm.
\end{abstract}

\section{Introduction}

Nearest-neighbor and weighted majority voting methods have been widely used
in medical image segmentation, originally at the pixel or voxel level
\cite{label_fusion_generative_model_2010}
and more recently for image patches
\cite{rueckert_patches_2013,coupe_2011,rousseau_2011,wachinger_2013}.
Perhaps the primary reason for the popularity of these nonparametric methods
is that standard label fusion techniques for image segmentation require robust
nonrigid registration whereas patch-based methods sidestep nonrigid image
alignment altogether. Thus, patch-based approaches provide a promising
alternative to registration-based methods for problems that present alignment
challenges, as in the case of whole body scans or other applications
characterized by large anatomical variability.

A second reason for patch-based methods' growing popularity lies in their
efficiency of computation: fast approximate nearest-neighbor search
algorithms, tailored for patches \cite{patchmatch} and for high-dimensional
spaces more generally
(e.g.,
\cite{ailon_2006_ann_jl,muja_flann_2009}),
can rapidly find similar patches, and can readily parallelize across
search queries. For problems where the end goal is segmentation or a
decision based on segmentation, solving numerous nonrigid registration
subproblems required for
standard label fusion could be a computationally expensive detour that, even if
successful, might not produce better solutions than a patch-based approach.

Many patch-based image segmentation methods can be viewed as variations of the
following simple algorithm. To determine whether a pixel in the new image
should be foreground (part of the object of interest) or background, we
consider the patch centered at that pixel. We compare this image patch to
patches in a training database, where each training patch is labeled either
foreground or background depending on the pixel at the center of the training
patch. We transfer the label from the closest patch in the training database
to the pixel of interest in the new image. A plethora of embellishments
improve this algorithm, such as, but not limited to, using $K$ nearest
neighbors or weighted majority voting instead of only the nearest neighbor
\cite{coupe_2011,rousseau_2011,wachinger_2013}, incorporating hand-engineered
or learned feature descriptors \cite{wachinger_2013}, cleverly choosing the
shape of a patch \cite{wachinger_2013}, and enforcing consistency among
adjacent pixels by assigning each training intensity image patch to a label
patch rather than a single label \cite{rousseau_2011,wachinger_2013}, or by
employing a Markov random field \cite{mrf_superresolution}.

Despite the broad popularity and success of nonparametric patch-based image
segmentation and the smorgasbord of tricks to enhance its performance, the
existing work has been empirical with no theoretical justification for when
and why such methods should work and, if so, how well and with how much
training data. In this paper, we bridge this gap between theory and practice
for nonparametric patch-based image segmentation algorithms. We propose a
probabilistic model for image segmentation that draws from recent work on
modeling natural imagery patches \cite{epll,patch_gmm}. We begin in Section
\ref{sec:myopic} with a simple case of our model that corresponds to inferring
each pixel's label separately from other pixels. For this special case of
so-called pointwise segmentation, we provide a theoretical performance
guarantee for patch-based nearest-neighbor and weighted majority voting
segmentation in terms of the available training data. Our analysis borrows
from existing theory on nonparametric time series classification
\cite{georgehc_2013_time_series_nips} and crucially relies on a new structural
property on neighboring patches being sufficiently similar. We present our
full model in Section \ref{sec:full} and derive a new iterative patch-based
image segmentation algorithm that combines ideas from patch-based image
restoration \cite{epll} and distributed optimization \cite{boyd_admm_2011}.
This algorithm alternates between inferring label patches separately and
merging these local estimates to form a globally consistent segmentation. We
show how various existing patch-based algorithms are special cases of this new
algorithm.

\section{Pointwise Segmentation and a Theoretical Guarantee\label{sec:myopic}}

For an image $A$, we use $A(i)$ to denote the value of image~$A$ at pixel $i$,
and $A[i]$ to denote the patch of image $A$ centered at pixel $i$ based on a
pre-specified patch shape; $A[i]$ can include feature descriptors in addition
to raw intensity values. Each pixel $i$ belongs to a finite, uniformly sampled
lattice~$\mathcal{I}$.

\textbf{Model.}
Given an intensity image $Y$, we infer its label image~$L$ that delineates an
object of interest in $Y$. In particular, for each pixel $i \in \mathcal{I}$,
we infer label $L(i) \in \{+1, -1\}$, where $+1$ corresponds to foreground
(object of interest) \mbox{and $-1$} to background. We make this inference
using patches of image $Y$, each patch of dimensionality~$d$ (e.g., for 2D
images and 5-by-5 patches, \mbox{$d=5^2=25$}). We model the joint distribution
$p(L(i),Y[i])$ of label $L(i)$ and image patch $Y[i]$ as a generalization of a
Gaussian mixture model (GMM) with diagonal \mbox{covariances, where} each
mixture component corresponds to either \mbox{$L(i)=+1$} or \mbox{$L(i)=-1$}.
We define this generalization, called a \textit{diagonal sub-Gaussian mixture
model}, shortly.

First, we provide a concrete example where label $L(i)$ and patch $Y[i]$ are
related through a GMM with $\mathcal{C}_i$ mixture components. Mixture
component $c \in \{1, \dots, \mathcal{C}_i\}$ occurs with probability
$\rho_{ic} \in [0,1]$ and has mean vector $\mu_{ic} \in \mathbb{R}^d$ and
label $\lambda_{ic} \in \{+1,-1\}$. In this example, we assume that all
covariance matrices are $\sigma^2 \mathbf{I}_{d\times d}$, and that there
exists constant $\rho_{\min}>0$ such that $\rho_{ic} \ge \rho_{\min}$ for all
$i,c$. Thus, image patch $Y[i]$ belongs to mixture component $c$ with
probability $\rho_{ic}$, in which case $Y[i] = \mu_{ic} + W_i$, where vector
$W_i \in \mathbb{R}^d$ consists of white Gaussian noise with variance
$\sigma^2$, and $L(i)=\lambda_{ic}$. Formally,
\[
p(L(i),Y[i])
=\sum_{c=1}^{\mathcal{C}_i}
   \rho_{ic}~\!
   \mathcal{N}(Y[i];\mu_{ic},\sigma^2 \mathbf{I}_{d\times d})
   \delta(L(i)=\lambda_{ic}),
\]
where $\mathcal{N}(\cdot; \mu, \Sigma)$ is a Gaussian density with mean $\mu$
and covariance $\Sigma$, and $\delta(\cdot)$ is the indicator function. 

The \textit{diagonal sub-Gaussian mixture model} refers to a 
generalization where noise vector $W_i$ consists of zero-mean i.i.d.~random
entries whose distribution has tails that decay at least as fast as that of a
Gaussian random variable. Formally, a zero-mean random variable $X$ is
sub-Gaussian 
with parameter $\sigma$ if its moment
generating function $\mathbb{E}[e^{s X}]$ satisfies $\mathbb{E}[e^{s X}]
\le e^{s^2\sigma^2/2}$ for all $s \in \mathbb{R}$. Examples of such random
variables include
$\mathcal{N}(0,\sigma^2)$ and $\text{Uniform}[-\sigma,\sigma]$.

Every pixel is associated with its own diagonal sub-Gaussian mixture model
whose parameters
$(\rho_{ic}, \mu_{ic}, \lambda_{ic})$ for $c = 1,\dots,\mathcal{C}_i$
are fixed but unknown. Similar to recent work on modeling
generic natural image patches \cite{epll,patch_gmm}, we do not model how
different overlapping patches behave jointly and instead only model how each
individual patch, viewed alone, behaves. We suspect that medical image patches
have even more structure than generic natural image patches, which are very
accurately modeled by a GMM~\cite{patch_gmm}.

Rather than learning the mixture model components, we instead take a
nonparametric approach, using available training data in nearest-neighbor or
weighted majority voting schemes to infer label $L(i)$ from image patch
$Y[i]$. To this end, we assume we have access to~$n$ i.i.d.~training
intensity-label \textit{image} pairs $(Y_1,L_1), \dots, (Y_n,L_n)$ that obey
our probabilistic model above.

\textbf{Inference.}
We consider two simple segmentation methods that operate on each pixel $i$
separately, inferring label $L(i)$ only based on image patch $Y[i]$.

\textit{Pointwise nearest-neighbor segmentation} first finds which training
intensity image $Y_u$ has a patch centered at pixel $j$ that is closest
to observed patch $Y[i]$. This amounts to solving
$
(\widehat{u},\widehat{j})
=\text{argmin}_{u\in\{1,2,\dots,n\}, j \in N(i)}
    \|Y_u[j]-Y[i]\|^2,
$
where $\|\cdot\|$ denotes Euclidean norm, and $N(i)$ refers to a
user-specified finite set of pixels that are neighbors of pixel $i$. Label
$L(i)$ is estimated to be $L_{\widehat{u}}(\,\widehat{j}\,)$.

\textit{Pointwise weighted majority voting segmentation} first computes the
following weighted votes for labels $\ell \in \{+1, -1\}$:
\[
V_{\ell}(i|Y[i]; \theta)
\triangleq
  \sum_{u=1}^n
    \sum_{j\in N(i)}
      \exp\big(-\theta\|Y_u[j]-Y[i]\|^2\big)
      \delta(L_u(j)= \ell),
\]
where $\theta$ is a scale parameter, and $N(i)$ again refers to
user-specified neighboring pixels of pixel $i$. Label $L(i)$ is estimated to
be the label $\ell$ with the higher vote $V_\ell(i|Y[i];\theta)$.
Pointwise nearest-neighbor segmentation can be viewed as this weighted voting
with $\theta \rightarrow \infty$. Pointwise weighted majority voting
has been used extensively for patch-based segmentation
\cite{rueckert_patches_2013,coupe_2011,rousseau_2011,wachinger_2013}, where we
note that our formulation readily allows for one to choose which training
image patches are considered neighbors, what the patch shape is, and whether
feature descriptors are part of the intensity patch vector $Y[i]$. 

\textbf{Theoretical guarantee.}
The model above allows nearby pixels to be associated with dramatically
different mixture models. However, real images are ``smooth'' with patches
centered at two adjacent pixels likely similar. We incorporate this smoothness
via a structural property on the sub-Gaussian mixture model parameters
associated with nearby pixels.
%
We refer to this property as the \textit{jigsaw condition},
which holds
if for every mixture component
$(\rho_{ic},\mu_{ic},\lambda_{ic})$ of the diagonal sub-Gaussian mixture model
associated with pixel $i$, there exists a neighbor $j \in N^*(i)$ such that
the diagonal sub-Gaussian mixture model associated with pixel $j$ also has a
mixture component with mean $\mu_{ic}$, label $\lambda_{ic}$, and mixture
weight at least $\rho_{\min}$; this weight need not be equal to $\rho_{ic}$.
The shape and size of neighborhood $N^*(i)$, which is fixed and unknown,
control how similar the mixture models are across image pixels.
%
Note that $N^*(i)$ affects how far from pixel $i$ we should look for training
patches, i.e., how to choose neighborhood $N(i)$ in pointwise nearest-neighbor
and weighted majority voting segmentation, where ideally $N(i) = N^*(i)$.

\emph{Separation gap.}
Our theoretical result also depends on the separation ``gap'' between training
intensity image patches that correspond to the two different labels:
\[
\mathcal{G}
\triangleq\min_{\substack{u,v\in\{1,\dots,n\},\\
i\in\mathcal{I},j\in N(i)\text{ s.t. }L_u(i)\ne L_v(j)}
}\|Y_u[i]-Y_v[j]\|^2.
\]
Intuitively, a small separation gap corresponds to the case of two training
intensity image patches that are very similar but one corresponds to
foreground and the other to background. In this case, a nearest-neighbor
approach may easily select a patch with the wrong label, resulting in an
error.

We now state our main theoretical guarantee. The proof is left to the
supplementary material and builds on existing time series classification
analysis~\cite{georgehc_2013_time_series_nips}.

\begin{theorem}
\label{thm:1nn-wmv-patches}
Under the model above with $n$ training intensity-label image pairs and
provided that the jigsaw condition holds with neighborhood $N^*$, pointwise
nearest-neighbor and weighted majority voting segmentation (with
user-specified neighborhood $N$ satisfying $N(i)\supseteq N^*(i)$ for every pixel $i$ and
with parameter $\theta = \frac{1}{8\sigma^2}$ for weighted majority
voting) achieve expected pixel labeling error rate
\begin{align*}
\mathbb{E}\bigg[
  \frac{1}{|\mathcal{I}|}
    \sum_{i\in\mathcal{I}}
      \delta(\text{mislabel pixel }i)
\bigg]
&\le
   |\mathcal{I}|\mathcal{C}_{\max}\exp\Big(-\frac{n\rho_{\min}}{8}\Big)
   +
   |N|n\exp\Big(-\frac{\mathcal{G}}{16\sigma^{2}}\Big),
\end{align*}
where
$\mathcal{C}_{\max}$ is the maximum number of mixture components of any
diagonal sub-Gaussian mixture model associated with a pixel, and
$|N|$ is the largest user-specified neighborhood of any pixel.
\end{theorem}

\noindent
To interpret this theorem, we consider sufficient conditions for each term on
the right-hand side bound to be at most $\varepsilon/2$ for $\varepsilon \in
(0,1)$. For the first term, the number of training intensity-label image pairs
$n$ should be sufficiently large so that we see all the different mixture model
components in our training data:
$
n
\ge
  \frac{8}{\rho_{\min}}
  \log( 2|\mathcal{I}|\mathcal{C}_{\max}/\varepsilon )
$.
For the second term, the gap $\mathcal{G}$ should be sufficiently large so
that the nearest training intensity image patch found does not produce a
segmentation error:
$\mathcal{G}
\ge
  16\sigma^2
  \log( 2|N|n/\varepsilon )$.
There are different ways to change the gap, such as changing the patch shape
and including hand-engineered or learned patch features. 
Intuitively, using
larger patches $d$ should widen the gap, but using larger patches also means
that the maximum number of mixture components $\mathcal{C}_{\max}$ needed to
represent a patch increases, possibly quite dramatically. 

The dependence on $n$ in the second term results from a worst-case analysis.
To keep the gap from having to grow as $\log(n)$, we could subsample the
training data so that $n$ is large enough to capture the diversity of mixture
model components yet not so large that it overcomes the gap. In particular,
treating $\mathcal{C}_{\max}$, $\sigma^2$, and $\rho_{\min}$ as constants that
depend on the application of interest and could potentially be estimated from
data, collecting $n = \Theta(\log(|\mathcal{I}|/\varepsilon))$ training image
pairs and with a gap
$\mathcal{G}
= \Omega\big(\log(
               (|N|\log |\mathcal{I}|)/\varepsilon
             )\big)$,
both algorithms achieve an expected error rate of at most $\varepsilon$. The
intuition is that as $n$ grows large, if we continue to consider all training
subjects, even those that look very different from the new subject, we are
bound to get unlucky (due to noise in intensity images) and, in the worst
case, encounter a training image patch that is close to a test image patch but
has the wrong label. Effectively, outliers in training data muddle
nearest-neighbor inference, and more training data means possibly more
outliers.

\section{Multipoint Segmentation}
\label{sec:full}

\textbf{Model.}
We generalize the basic model to infer label patches $L[i]$ rather than just a
single pixel's label $L(i)$. Every label patch $L[i]$ is assumed to have
dimensionality $d'$, where $d$ and $d'$ need not be equal. For example, for 2D
images, $Y[i]$ could be a 5-by-5 patch ($d=25$) whereas $L[i]$ could be a
3-by-3 patch ($d'=9$). When $d'>1$, estimated label patches must be merged to
arrive at a globally consistent estimate of label image~$L$. This case is
referred to as multipoint segmentation.

In this general case, we assume there to be $k$ latent label images
$\Lambda_1, \dots, \Lambda_k$ that occur with probabilities $\pi_1, \dots,
\pi_k$. To generate intensity image $Y$, we first sample label
image $\Lambda \in \{\Lambda_1, \dots, \Lambda_k\}$ according to probabilities
$\pi_1, \dots, \pi_k$. Then we sample label image $L$ to be a perturbed
version of $\Lambda$ such that
$
p(L|\Lambda) \propto \exp(-\alpha\mathbf{d}(L, \Lambda))
$
for some constant $\alpha \ge 0$ and differentiable ``distance'' function
$\mathbf{d}(\cdot, \cdot)$. For example, $\mathbf{d}(L, \Lambda)$ could relate
to volume overlap between the segmentations represented by label images $L$
and $\Lambda$ with perfect overlap yielding distance 0. Finally, intensity
image~$Y$ is generated so that for each pixel $i\in\mathcal{I}$, patch $Y[i]$
is a sample from a mixture model patch prior $p(Y[i]|L[i])$. If $\alpha=0$,
$d'=1$, and the mixture model is diagonal sub-Gaussian, we get our
earlier~model.

We refer to this formulation as a \textit{latent source model} since the
intensity image patches could be thought of as generated from the latent
``canonical'' label images $\Lambda_1, \dots, \Lambda_k$ combined with the
latent mixture model clusters linking label patches $L[i]$ to intensity
patches $Y[i]$. This hierarchical structure enables local appearances around a
given pixel to be shared across the canonical label~images.

\textbf{Inference.}
We outline an iterative algorithm based on the expected patch log-likelihood
(EPLL) framework \cite{epll}, deferring details to the supplementary material.
The EPLL framework seeks a label image $L$ by solving
\begin{equation*}
\widehat{L}
=\!\!
\underset{L \in \{+1,-1\}^{|\mathcal{I}|}}
         {\text{argmax}}
  \bigg\{\!
    \log\bigg(
          \sum_{g=1}^k
            \pi_g
            \exp(-\alpha\mathbf{d}(L,\Lambda_g))
        \bigg)
    +
    \frac{1}{|\mathcal{I}|}
    \sum_{i\in\mathcal{I}}\log p(Y[i]|L[i])\!
  \bigg\}.
\end{equation*}
The first term in the objective function encourages label image $L$ to be
close to the true label images $\Lambda_1, \dots, \Lambda_k$. The second term
is the ``expected patch log-likelihood'', which favors solutions whose local
label patches agree well on average with the local intensity patches according
to the patch priors. Since latent label images $\Lambda_1, \dots, \Lambda_k$
are unknown, we use training label images $L_1, \dots, L_n$ as proxies
instead, replacing the first term in the objective function with
$F(L;\alpha)
 \triangleq
 \log
 \big(
 \frac{1}{n}
 \sum_{u=1}^n
   \exp(-\alpha\mathbf{d}(L,L_u))\big)$.
Next, we approximate the unknown patch prior $p(Y[i]|L[i])$ with a kernel
density estimate
\[
\widetilde{p}(Y[i]|L[i];\gamma)
\propto\sum_{u=1}^{n}\sum_{j \in N(i)}
\mathcal{N}\Big(
             Y[i];
             Y_u[j],
             \frac{1}{2\gamma}\mathbf{I}_{d\times d}
             \Big)
             \, 
             \delta(L[i]=L_u[j]),
\]
where the user specifies a neighborhood $N(i)$ of pixel $i$, and constant
$\gamma>0$ that controls the Gaussian kernel's bandwidth. We group the pixels
so that nearby pixels within a small block all share the same kernel density
estimate. This approximation assumes a stronger version of the
jigsaw condition from Section \ref{sec:myopic} since the algorithm operates as
if nearby pixels have the same mixture model as a patch prior. We thus
maximize objective
$F(L;\alpha) +
 \frac{1}{|\mathcal{I}|}
 \sum_{i \in \mathcal{I}} \log \widetilde{p}(Y[i]|L[i];\gamma)$. 

Similar to the original EPLL method~\cite{epll}, we introduce an auxiliary
variable $\xi_{i}\in\mathbb{R}^{d'}$ for each patch $L[i]$, where $\xi_i$ acts
as a local estimate for $L[i]$. Whereas two patches $L[i]$ and $L[j]$ that
overlap in label image~$L$ must be consistent across the overlapping pixels,
there is no such requirement on their local estimates $\xi_i$ and $\xi_j$. In
summary, we maximize the objective
$
F(L;\alpha)+
\frac{1}{|\mathcal{I}|}
\sum_{i \in \mathcal{I}} \log \widetilde{p}(Y[i]|\xi_i;\gamma)
-\frac{\beta}{2}\sum_{i \in \mathcal{I}} \|L[i] - \xi_i\|^2
$ for $\beta>0$, subject to constraints $L[i] = \xi_i$ that are enforced
using Lagrange multipliers. We numerically optimize this cost function using the
Alternating Direction Method of Multipliers for
distributed optimization \cite{boyd_admm_2011}. Given the current estimate of
label image $L$, the algorithm produces estimate $\xi_i$ for $L[i]$ given
$Y[i]$ in parallel across $i$. Next, it updates $L$ based on $\xi_i$ via a
gradient method. Finally, the Lagrange multipliers are updated to penalize
large discrepancies between $\xi_i$ and $L[i]$.

Fixing $\xi_i$ and updating $L$ corresponds to merging local patch estimates
to form a globally consistent segmentation. This is the only step that
involves expression $F(L;\alpha)$. With $\alpha=0$ and forcing the Lagrange
multipliers to always be zero, the merging becomes a simple averaging of
overlapping label patch estimates $\xi_i$. This algorithm corresponds to
existing multipoint patch-based segmentation algorithms
\cite{coupe_2011,rousseau_2011,wachinger_2013} and the in-painting technique
achieved by the original EPLL method. Setting $\alpha=\beta=0$ and $d'=1$
yields pointwise weighted majority voting with parameter $\theta = \gamma$.
When $\alpha > 0$, a global correction is applied, shifting the label image
estimate closer to the training label images. This should produce better
estimates when the full training label images can, with small perturbations
as measured by $\mathbf{d}(\cdot, \cdot)$, explain new intensity images.

 
\begin{wrapfigure}[14]{r}[0pt]{3.7cm}
\includegraphics[height=3.5cm,clip=true,trim=1em 1.6em 0 1.4em]{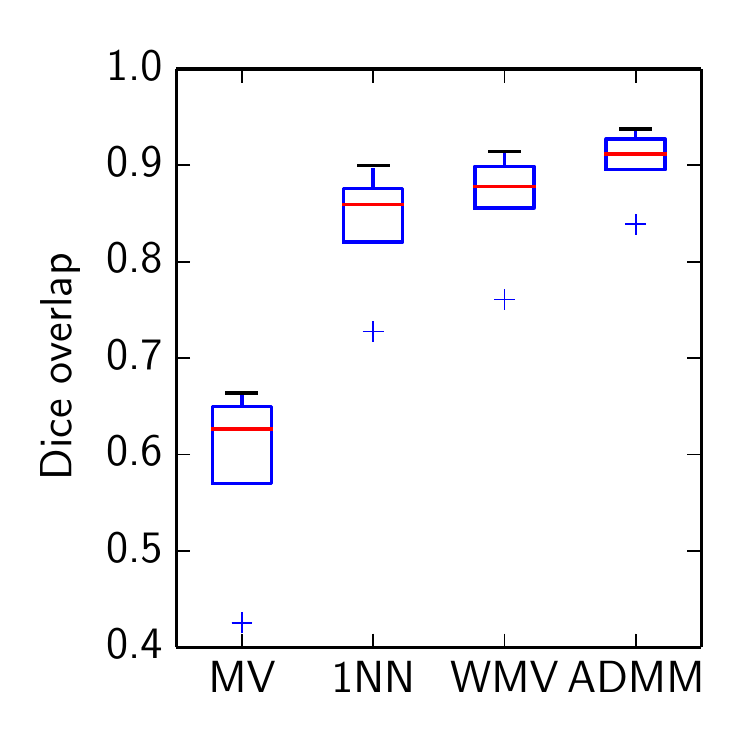}
\caption{Liver segmentation results.}
\label{fig:dice}
\end{wrapfigure}

\textbf{Experimental results.}
We empirically explore the new iterative algorithm on 20 labeled
thoracic-abdominal contrast-enhanced CT scans from the Visceral
\textsc{anatomy3} dataset \cite{anatomy3}. We train the model on 15 scans and
test on the remaining 5 scans. The training procedure amounted to using 10 of
the 15 training scans to estimate the algorithm parameters in an exhaustive
sweep, using the rest of the training scans to evaluate parameter settings.
Finally, the entire training dataset of 15 scans is used to segment the test
dataset of 5 scans using the best parameters found during training. For each
test scan, we first use a fast affine registration to roughly align each
training scan to the test scan. Then we apply four different algorithms: a
baseline majority voting algorithm (denoted ``MV") that simply averages the
training label images that are now roughly aligned to the test scan, pointwise
nearest neighbor (denoted ``1NN") and weighted majority voting (denoted
``WMV") segmentation that both use approximate nearest patches, and finally
our proposed iterative algorithm (denoted ``ADMM''), setting distance
$\mathbf{d}$ to one minus Dice overlap. Note that Dice overlap can be
reduced to a differentiable function by relaxing our optimization to allow
each label to take on a value in $[-1, 1]$. By doing so, the Dice overlap of
label images $L$ and $\Lambda$ is given by
$2\langle \widetilde{L}, \widetilde{\Lambda}\rangle/
(\langle \widetilde{L}, \widetilde{L} \rangle +
 \langle \widetilde{\Lambda}, \widetilde{\Lambda} \rangle)$,
where $\widetilde{L} = (L+1)/2$ and $\widetilde{\Lambda} = (\Lambda+1)/2$.

We segmented the liver, spleen, left kidney, and right kidney. We report Dice
overlap scores for the liver in Fig.~\ref{fig:dice} using the four algorithms.
Our results for segmenting the other organs follow a similar trend where the
proposed algorithm outperforms pointwise weighted majority voting, which
outperforms both pointwise nearest-neighbor segmentation and the baseline
majority voting. For the organs segmented, there was little benefit to having
$\alpha>0$, suggesting the local patch estimates to already be quite
consistent and require no global correction.

\section{Conclusions}

We have established a new theoretical performance guarantee for two
nonparametric patch-based segmentation algorithms, uniting recent lines of
work on modeling patches in natural imagery and on theory for nonparametric
time series classification. Our result indicates that if nearby patches behave
as mixture models with sufficient similarity, then a myopic segmentation works
well, where its quality is stated in terms of the available training data.
Our main performance bound provides insight into how one should
approach building a training dataset for patch-based segmentation.
The looseness in the bound could be attributed to outliers in training data.
Detecting and removing these outliers should lead to improved segmentation
performance.


From a modeling standpoint, understanding the joint behavior of patches could
yield substantial new insights into exploiting macroscopic structure in images
rather than relying only on local properties. In a related direction, while we
have modeled the individual behavior of patches, an interesting theoretical
problem is to find joint distributions on image pixels that lead to such
marginal distributions on patches. Do such joint distributions exist? If not,
is there a joint distribution whose patch marginals approximate the mixture
models we use? These questions outline rich areas for future research.

\smallskip{}
\noindent
\textbf{Acknowledgments.} This work was supported in part by the NIH NIBIB
NAC P41EB015902 grant, MIT Lincoln Lab, and Wistron Corporation.
GHC was supported by an NDSEG fellowship.

\bibliographystyle{plain}
\bibliography{latent_source_short}

\end{document}